\documentclass[manuscript]{acmart}

\usepackage{multirow}
\usepackage{makecell}
\usepackage{caption}
\usepackage{subcaption}
\usepackage{graphicx}
\usepackage{amsmath}
\usepackage{tabularx}
\DeclareMathOperator*{\argmin}{argmin} 

\AtBeginDocument{%
  \providecommand\BibTeX{{%
    \normalfont B\kern-0.5em{\scshape i\kern-0.25em b}\kern-0.8em\TeX}}}

\copyrightyear{2022}
\acmYear{2022}
\setcopyright{acmlicensed}
\acmConference[HT '22]{Proceedings of the 33rd ACM Conference on Hypertext and Social Media}{June 28-July 1, 2022}{Barcelona, Spain}
\acmBooktitle{Proceedings of the 33rd ACM Conference on Hypertext and Social Media (HT '22), June 28-July 1, 2022, Barcelona, Spain}
\acmPrice{15.00}
\acmDOI{10.1145/3511095.3531269}
\acmISBN{978-1-4503-9233-4/22/06}

\acmSubmissionID{1031}

\begin{document}

\title{Learning to Adapt Domain Shifts of Moral Values via Instance Weighting}


\author{Xiaolei Huang}
\authornotemark[1]
\orcid{0000-0003-0478-8715}
\affiliation{%
  \institution{University of Memphis}
  \city{Memphis}
  \state{Tennessee}
  \country{United States}}
\email{xiaolei.huang@memphis.edu}

\author{Alexandra Wormley}
\email{awormley@asu.edu}
\orcid{0000-0001-9892-6087}
\author{Adam Cohen}
\email{adam.cohen@asu.edu}
\affiliation{%
  \institution{Arizona State University}
  \city{Tempe}
  \state{Arizona}
  \country{United States}
}

\begin{abstract}
Classifying moral values in user-generated text from social media is critical in understanding community cultures and interpreting user behaviors of social movements. 
Moral values and language usage can change across the social movements; however, text classifiers are usually trained in source domains of existing social movements and tested in target domains of new social issues without considering the variations.
In this study, we examine domain shifts of moral values and language usage, quantify the effects of domain shifts on the morality classification task, and propose a neural adaptation framework via instance weighting to improve cross-domain classification tasks. 
The quantification analysis suggests a strong correlation between morality shifts, language usage, and classification performance.
We evaluate the neural adaptation framework on a public Twitter data across 7 social movements and gain classification improvements up to 12.1\%.
Finally, we release a new data of the COVID-19 vaccine labeled with moral values and evaluate our approach on the new target domain.
For the case study of the COVID-19 vaccine, our adaptation framework achieves up to 5.26\% improvements over neural baselines.
\textbf{This is the first study} to quantify impacts of moral shifts, propose adaptive framework to model the shifts, and conduct a case study to model COVID-19 vaccine-related behaviors from moral values.
\end{abstract}



\maketitle

\section{Introduction}

Moral values defined by the Moral Foundations Theory (MFT)~\cite{haidt2007new, graham2013moral} provide a unique perspective to interpret human behaviors in various social movements (e.g., gun control~\cite{brady2017emotion, roy2021analysis}) by connecting cultures and reflecting human ideology.
The theory summarizes moral ideologies with five opposing pairs (virtues and vices) of foundations, including authority/subversion, care/harm, fairness/cheating, loyalty/betrayal, and purity/degradation (Table~\ref{tab:moralt_values}).
Language usage in user-generated texts can reflect emotions, moral values, and cultures that explain behavior motivations and shape people's decision-making.
Emotional intuitions and moral values, known as ``foundation'' in the MFT, can influence people's judgments and decision-making.
Studies are leveraging online user-generated texts (e.g., tweets) at scale by building morality classifiers to examine online users' reactions and behaviors across different social movements, such as political stances~\cite{johnson2018classification, rezapour2021incorporating, reiter2021studying}, social protest sentiments~\cite{garten2016morality, mooijman2018moralization, xie2019text, rezapour2019enhancing}, and natural disaster attitudes~\cite{dawson2012morality, dickinson2016moral, wolsko2016red}.
However, limited classifiers have explicitly considered the shifts of moral values across social movements.

\begin{table*}[htp]
\centering
\begin{tabularx}{\textwidth}{c|c|X}
\multicolumn{2}{c|}{Moral Foundation} & \multirow{2}{*}{Description} \\
Virtue & Vice &  \\\hline\hline
authority & subversion & The values underlies virtues of leadership and followership, deference to legitimate authority and respect for traditions \\\hline
care & harm & This foundation describes ability to feel (and dislikes) pain of others. It underlies kindness, gentleness, and nurturance. \\\hline
fairness & cheating & The values generate ideas of justice, rights, and autonomy. \\\hline
loyalty & betrayal & The foundation underlies virtues of patriotism and self-sacrifice for the group. \\\hline
purity & degradation & The foundation underlies 1) an idea that the body is a temple that can be desecrated by immoral activities and contaminants, and 2) religious notions of striving to live in an elevated, less carnal, more noble way.
\end{tabularx}
\caption{Principles and descriptions of moral foundations.}
\label{tab:moralt_values}
\end{table*}

Shifts in language use and morality across social movements can impact the applications of well-trained morality classifiers.
Studies show that moral values vary across \textit{domains} of social issues, such as variations between \#ALLLivesMatter and \#BlackLivesMatter~\cite{rezapour2019how}, political polarization in immigration and gun control~\cite{roy2021analysis}, and moral differences between liberals and conservatives~\cite{stewart2021moving, reiter2021studying}.
The \textit{domain} is implicitly embedded in the classification process that may impact classifiers to assess user behaviors and ideologies: classifiers are often built to be applied to a new target domain that doesn't yet exist, and performance on held-out data is measured to estimate performance on the new domain whose distributions of moral values and language use may vary from existing domains of training data.
It is also hard to annotate moral values of a new domain corpus in the field of computational social sciences, which requires extensive human labor and professional training procedures~\cite{johnson2018classification, hoover2020moral}.
For example, a study~\cite{johnson2018classification} shows that demographics and political biases of annotators yield skewed annotations towards care/harm or fairness/cheating using the Amazon Mechanical Turk, which is a widely-used crowdsourcing platform to obtain data annotations.
The domain shift is a challenge to build robust classifiers when the domain of training data is different from the domain of testing data and obtaining high-quality annotations is hard.


In this study, we focus on the morality classification task to provide insights into how performance varies across social movement domains and propose a neural adaptation framework that applies instance weighting~\cite{jiang2007instance} to model the domain shift.
The instance weighting is a domain adaptation approach that re-weights the source-domain training instances to approximate the target-domain distribution.
We examine the changes on a publicly available data~\cite{hoover2020moral}, which contains tweets with moral value labels for 7 social issues (All Lives Matter, Baltimore Protest, Black Lives Matter, Hate Speech, US Presidential Election, MeToo movement, and Natural Disaster).
Our analysis considers domain shifts of language usage and morality annotations.
The study addresses the following research questions: 1) To what extent do the language use and morality vary across the social issues, 2) In what ways does the shift influence text classification across domains, and 3) Can morality classifiers be adapted to perform better in a new target domain.
To address question 1, we compare cross-domain distributions of language use and moral values and statistically examine relations between language and morality shifts.
To address question 2, we conduct cross-domain classification experiments that train a classifier on existing source domains and evaluate the classifier on a new target domain. 
Our analysis statistically quantifies the effects of domain shifts on classification performance by a two-tailed t-test.
To address question 3, we apply a standard instance weighting approach treating social issues as domains and adapting classifiers trained on existing source domains on a new target domain.
We show that this approach can significantly improve classification performance on the public dataset, even on a new social issue, the COVID-19 vaccine.

\section{Morality Data}
\label{sec:data}

\begin{table*}[ht]
\centering
\resizebox{1\textwidth}{!}{
    \begin{tabular}{c|c|c|ccccccccccc}
    Domain & Doc & Token & authority & betrayal & care & cheating & degradation & fairness & harm & loyalty & purity & subversion & no-moral \\ \hline\hline
    ALM & 3,389 & 16.76 & 6.37 & 1.04 & 12.07 & 13.22 & 3.16 & 13.51 & 19.03 & 6.29 & 2.14 & 2.35 & 20.82 \\
    Baltimore & 4,815 & 15.53 & 0.42 & 11.66 & 3.39 & 9.81 & 0.57 & 2.70 & 5.09 & 7.13 & 0.73 & 5.57 & 52.95 \\
    BLM & 4,980 & 15.41 & 5.31 & 2.81 & 6.09 & 14.06 & 4.32 & 8.49 & 19.74 & 7.73 & 2.84 & 5.91 & 22.7 \\
    Davidson & 4,641 & 15.64 & 0.41 & 0.82 & 0.19 & 1.26 & 1.36 & 0.08 & 2.74 & 0.78 & 0.08 & 0.12 & 92.16 \\
    Election & 4,994 & 17.87 & 2.65 & 2.00 & 6.27 & 9.53 & 2.09 & 8.79 & 9.19 & 3.28 & 6.40 & 2.56 & 47.23 \\
    MeToo & 4,313 & 23.50 & 7.17 & 5.29 & 3.53 & 10.93 & 14.83 & 6.55 & 7.07 & 5.42 & 3.08 & 14.89 & 21.23 \\
    Sandy & 3,847 & 14.49 & 9.60 & 3.11 & 21.41 & 9.9 & 1.93 & 3.74 & 16.83 & 8.93 & 1.45 & 9.73 & 13.38 \\\hline\hline
    Overall & 30,979 & 17.02 & 4.45 & 4.02 & 6.96 & 10.01 & 4.34 & 6.23 & 11.25 & 5.68 & 2.56 & 6.16 & 38.34
    \end{tabular}
}
\caption{Statistics summary of the multi-class morality dataset~\cite{hoover2020moral}. We list number of total document (Doc), average number of tokens per tweet document (Token), and label distributions of 11 annotations in percentage (\%).}
\label{tab:moral_data}
\end{table*}

In this study, we retrieved the publicly anonymized data~\cite{hoover2020moral} that have morality annotations for 35,108 tweets and 11 moral annotation types.
The tweets come from seven discourse domains that focus on major US social movements in the past few years: 1) ALM, tweets related to All Lives Matter movement that typically contains with conservative views and oppose to the ideas of the Black Lives Matter movement; 2) Baltimore, tweets related to the Baltimore protests against police brutality leading to the death of Freddie Gray; 3) BLM, tweets related to Black Lives Matter that fights against discrimination and inequality experienced by African American people; 4) Davidson, tweets of hate speech and offensive language collected by~\cite{davidson2017automated}; 5) Election, tweets posted during the 2016 U.S. presidential election; 6) MeToo, tweets related to the \#MeToo movement against sexual abuse and harassment; 7) Sandy, tweets related to the Hurricane Sandy, which was a natural disaster happened in 2012.
We replace any user mentions and URLs in the tweet documents with ``USER'' and ``URL'' to preserve user privacy.
We then lowercased all tweet documents and tokenized each document by NLTK~\cite{bird2004nltk}.
We removed tweet documents if they had fewer than 3 tokens.
The annotations have one no-moral and the other 10 moral annotations defined by the Moral Foundations Theory~\cite{haidt2007new, graham2013moral}.
The label set proposes five bipolar factor pairs (virtues / vices): care/harm, fairness/cheating, loyalty/betrayal, authority/subversion, purity/degradation.
We present descriptions of the moral foundations in Table~\ref{tab:moralt_values}.
The no-moral label indicates a tweet is nonmoral if the tweet does not fall under any foundations.
At least three annotators annotated each tweet document. 
To ensure annotation quality, we followed the previous study~\cite{hoover2020moral} and used the majority vote to keep any labels with votes by at least two annotators.
We dropped documents with empty annotations after applying the majority vote.
If most annotators label a tweet with the no-moral label, we mark the tweet as no-moral and drop other labels.

We summarize the data statistics in Table~\ref{tab:moral_data}.
The documents are short, with 17 tokens per document on average. 
However, we can find that skewed distributions exist in labels that the 11 document labels are imbalanced.
For example, the percentage of no-moral ranges from 13.38\% in the Sandy domain to 92.16\% in the Davidson domain.
The distributions of moral values are also different across the domains.
For example, the top 3 moral values of Election are cheating, harm, and fairness, but subversion, degradation, and cheating are the 3 top moral values for MeToo. 
To better understand moral variations across domains, we extract percentages of moral values for each domain and group the values using the Moral Foundations Theory into \textit{virtue} and \textit{vice} moral values.
We group the social events according to their virtue-vice ratio and summarize the distributions of moral values in Table~\ref{tab:moral_labels}.

\begin{table*}[htp]
\centering
\resizebox{1\textwidth}{!}{
    \begin{tabular}{c|ccccc|ccccc|c}
    \multirow{2}{*}{Domain} & \multicolumn{5}{c|}{Virtue (\%)} & \multicolumn{5}{c|}{Vice (\%)} & \multirow{2}{*}{\begin{tabular}[c]{@{}c@{}}Virtue-Vice\\ Ratio\end{tabular}} \\
     & authority & care & fairness & loyalty & purity & betrayal & cheating & degradation & harm & subversion &  \\\hline\hline
    Sandy & 11.08 & 24.72 & 4.31 & 10.30 & 1.68 & 3.58 & 11.43 & 2.23 & 19.43 & 11.23 & 1.09 \\
    Election & 5.03 & 11.88 & 16.67 & 6.21 & 12.12 & 3.79 & 18.06 & 3.97 & 17.42 & 4.85 & 1.08 \\
    ALM & 8.04 & 15.24 & 17.07 & 7.94 & 2.70 & 1.32 & 16.69 & 3.99 & 24.03 & 2.97 & 1.04 \\\hline
    BLM & 6.87 & 7.88 & 10.99 & 10.01 & 3.67 & 3.63 & 18.19 & 5.58 & 25.54 & 7.64 & 0.65 \\
    MeToo & 9.10 & 4.48 & 8.31 & 6.88 & 3.91 & 6.72 & 13.88 & 18.83 & 8.98 & 18.91 & 0.49 \\
    Baltimore & 0.88 & 7.21 & 5.74 & 15.15 & 1.54 & 24.78 & 20.85 & 1.21 & 10.81 & 11.84 & 0.44 \\
    Davidson & 5.25 & 2.36 & 1.05 & 9.97 & 1.05 & 10.50 & 16.01 & 17.32 & 34.91 & 1.57 & 0.25 \\\hline\hline
    Overall & 7.22 & 11.28 & 10.10 & 9.21 & 4.15 & 6.52 & 16.23 & 7.04 & 18.25 & 9.99 & 0.72
    \end{tabular}
}
\caption{Distributions of moral values across the 7 social domains. We group moral values into two groups, virtue and vice. The virtue-vice ratio is a divide operation between virtue-related and vice-related moral value counts. Overall, the data shows higher ratio of the vice-related moral values.}
\label{tab:moral_labels}
\end{table*}

Significant distribution variations of moral labels exist across 7 domains in Table~\ref{tab:moral_labels}.
For example, the three social protests (BLM, MeToo, and Baltimore) and the hate speech (Davidson) have shown very different moral value distributions than the other domains, such as presidential election (Election) and natural disaster (Sandy).
The ratio comparison between virtue-related and vice-related moral values can reflect that the social movements have higher vice-related moralities in their language.
Studies have shown that group identities relating to their cultural and social backgrounds can lead to moral variations~\cite{van2012conviction}.
For example, ALM and BLM show various moral value distributions in virtue-related and vice-related moralities; degradation and subversion are the top 2 moral values in the MeToo domain, while cheating and harm are the top 2 moral values in the Election domain. Our finding generalizes and aligns with a recent study~\cite{rezapour2019how} that moral values in ALM- and BLM-related social events are different. 
The moral variations motivate us to characterize effects and examine the impacts of moral variations.
We conducted the following analyses, moral value shifts and moral shift impacts, to understand two important questions to quantify the effects and impacts:
\begin{itemize}
    \item To what extent do the moral values shift across domains?
    \item To what extent can the moral shift impact morality modeling?
\end{itemize}

\subsection{Analysis 1: Moral Value Shifts}
\label{subsec:analysis1}
\begin{figure}[ht]
\centering
     \begin{subfigure}[ht]{0.47\textwidth}
         \includegraphics[width=\textwidth]{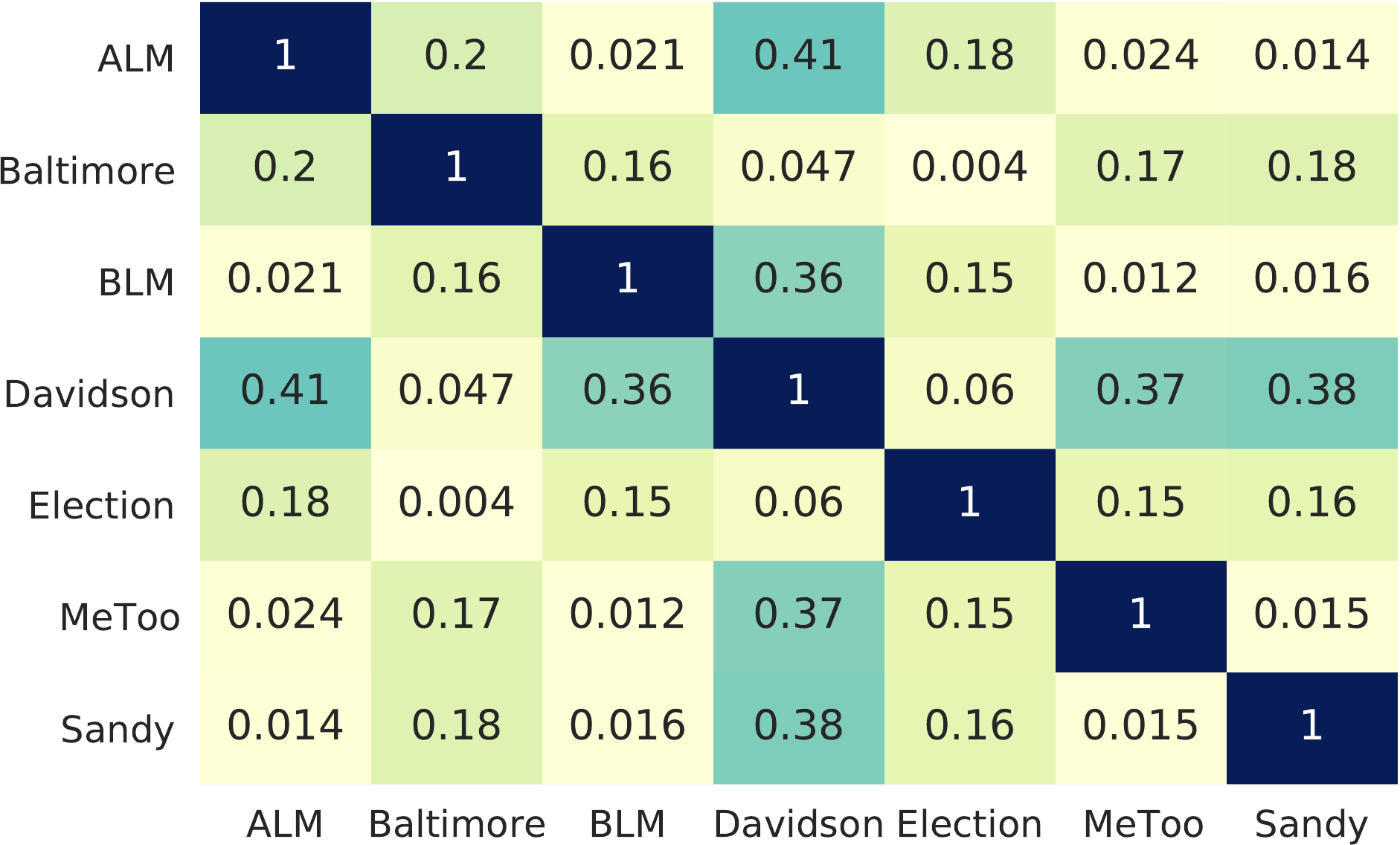}
         \caption{Similarities of moral values across domains.}
         \label{fig:label_sims}
     \end{subfigure}
     \hfill
     \begin{subfigure}[ht]{0.47\textwidth}
         \includegraphics[width=\textwidth]{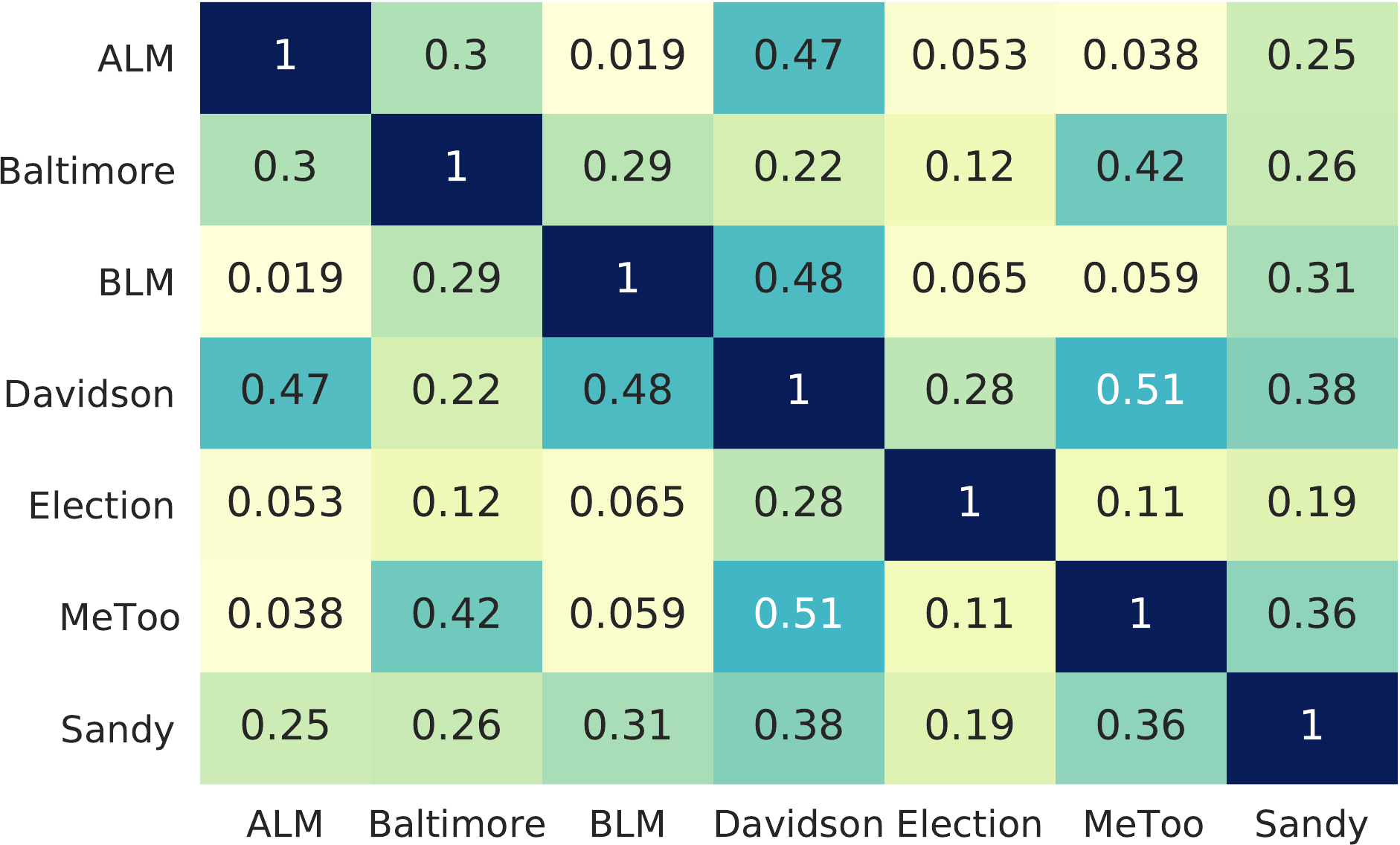}
         \caption{Similarities of topic distributions across domains.}
         \label{fig:topic_sims}
     \end{subfigure}

\caption{Analyses to quantify effects of moral shifts (Section~\ref{subsec:analysis1}). Higher scores indicate domains are more similar.}
\label{fig:analysis1}
\end{figure}

To quantify moral shifts, we measure cosine similarities of moral values and language usage.
Language usage is an important indicator to reflect and identify moral values~\cite{garten2016morality, johnson2018classification, rezapour2021incorporating, roy2021identifying}.
We extract linguistic patterns of language usage by a Latent Dirichlet Allocation (LDA)~\cite{blei2003latent}, which abstracts language usage into thematic vectors.
We train a \texttt{LdaModel} from GenSim~\cite{rehurek2010software} with 200 topics and default parameters over all available corpus. 
The model learns a multinomial topic distribution $P(Z|D)$ from a Dirichlet prior, where $Z$ refers to each topic, and $D$ refers to each document.
After training the topic model, we can encode each document $d$ with a probability distribution over the 200 topics. 
For each domain (e.g., BLM or MeToo), we calculate the average topic distribution across the documents from that domain.
Finally, we compute cosine similarities of the topic distributions between every two domains.
To measure morality similarities across domains, we reuse the moral label distributions (Table~\ref{tab:moral_labels}) and calculate cosine similarities between every two domains of the moral labels.
Figure~\ref{fig:label_sims} and \ref{fig:topic_sims} illustrate the cosine similarities across domains for morality (11 values) and language usage (topic) respectively.

We can find that similarities are low and varying across domains.
The low similarities indicate both language and moral values shift significantly across domains.
For example, similarities between ALM and BLM in moral values and language usage are 0.021 and 0.019, respectively.
The Davidson has comparatively high similarities from other domains.
We infer this as the domain includes diverse topics related to hate speech and offensive language.

We conduct a significance test to verify if the language and moral values variations correlate. 
The significance test requires the normality test to decide whether to use parametric or non-parametric methods.
We conduct a normality test via \texttt{normaltest} from \texttt{statsmodel}~\cite{seabold2010statsmodels}.
The test yields p-values as 0.00873 and 3.318e-05 for language and moral value variations respectively, which demonstrates that the significance test will use non-parametric methods.
We use the \texttt{spearmanr} from \texttt{statsmodel} to test our null hypothesis that the moral variations have no significant correlation with the topic variations. 
Finally, the results show p-value=$0.488$ and coefficient=$0.165$ for the test between domain variations of the label and linguistic expression.
Therefore, we can not reject the null hypothesis of the significant test.
The observations suggest that: while we can observe variations exist in both moral labels and language use,  the two variations across domains do have a strong correlation.

\subsection{Analysis 2: Moral Shift Impacts}
\label{subsec:analysis2}

\begin{figure}[ht]
\centering
     \begin{subfigure}[ht]{0.47\textwidth}
        \includegraphics[width=\textwidth]{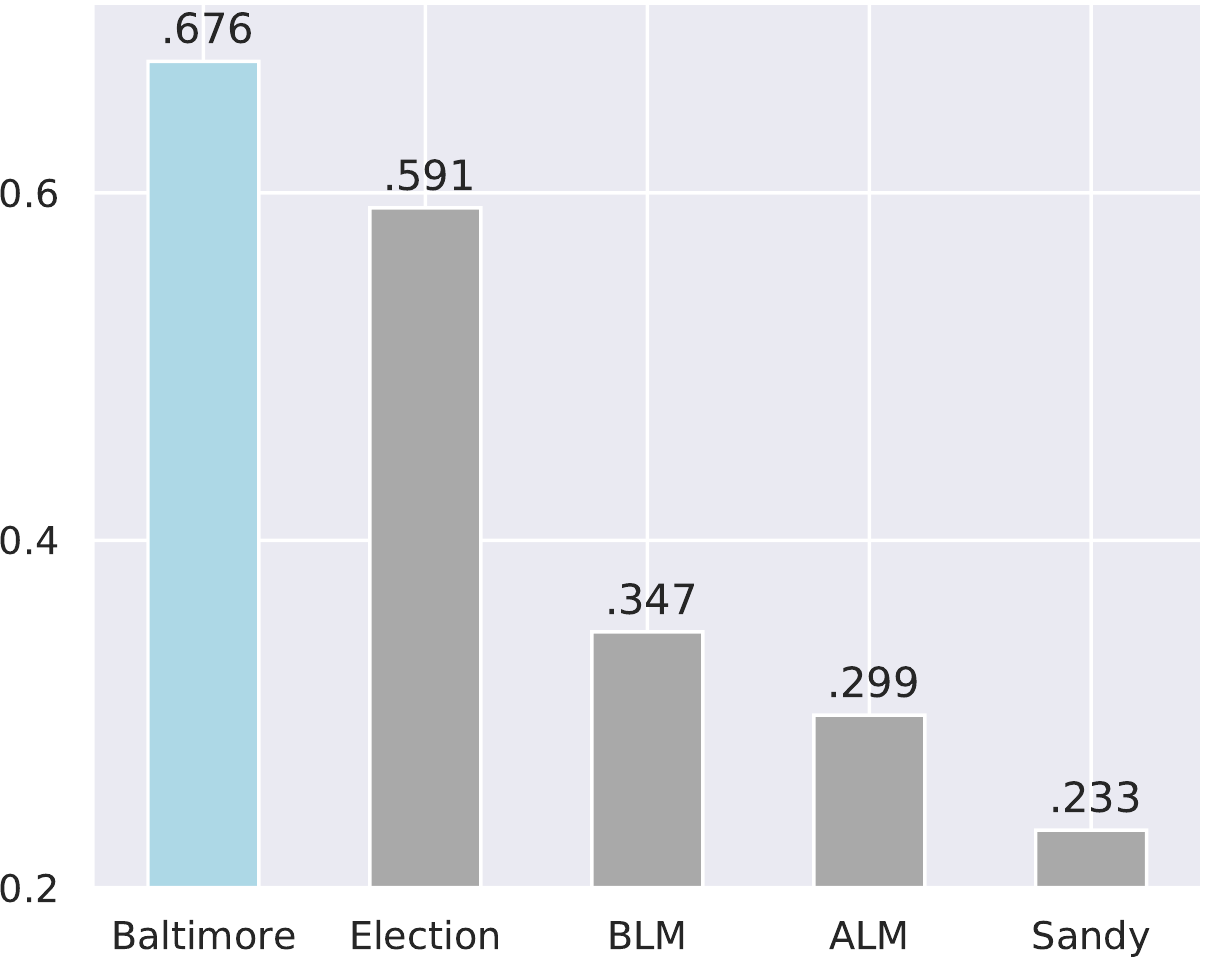}
        \caption{In-domain (Baltimore in light blue) vs. Cross-domain (others in grey) classification performance.}
        \label{fig:cross_baltimore}
    \end{subfigure}
    \hfill
    \begin{subfigure}[ht]{0.47\textwidth}
        \includegraphics[width=\textwidth]{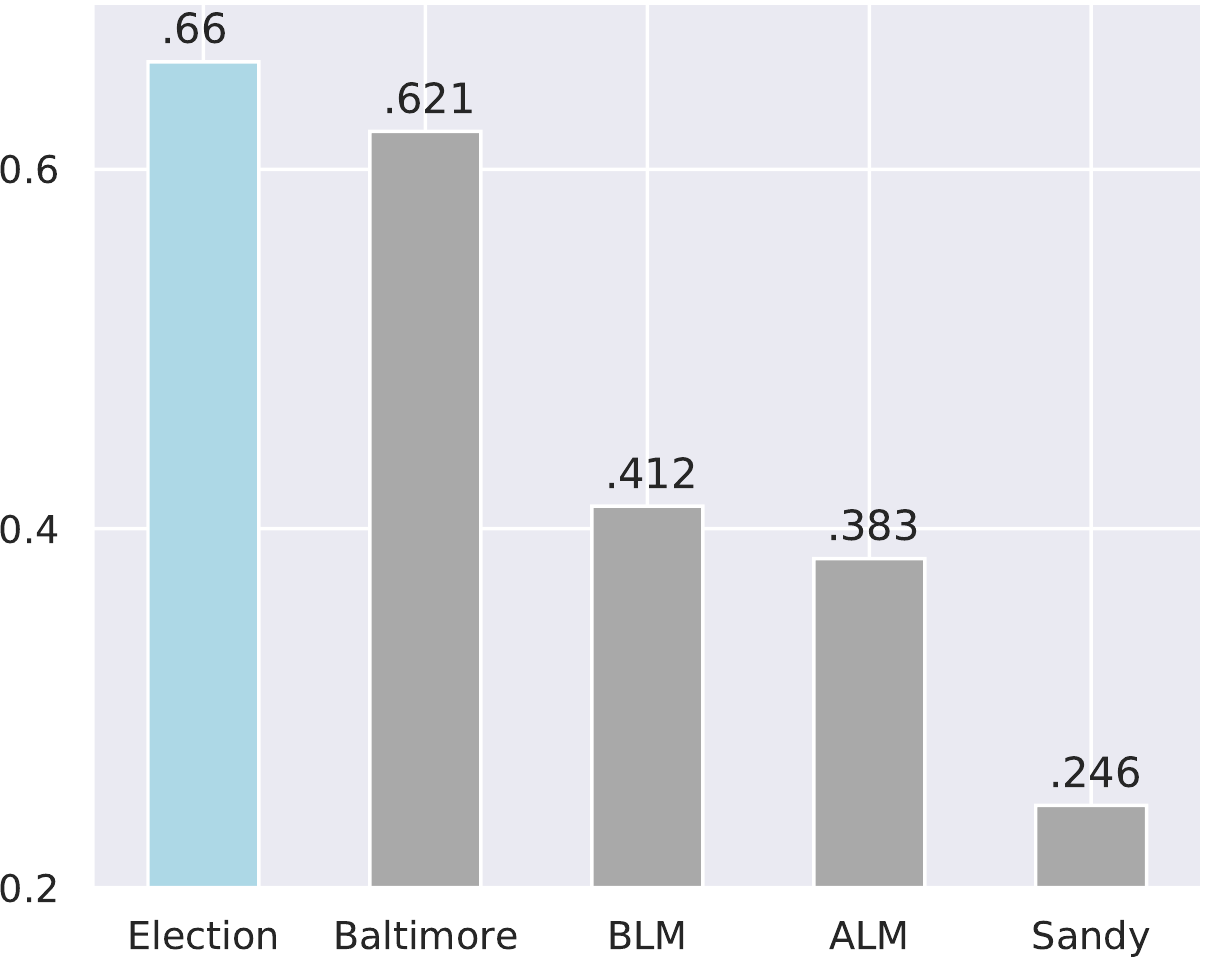}
        \caption{In-domain (Election in light blue) vs. Cross-domain (others in grey) classification performance.}
        \label{fig:cross_election}
    \end{subfigure}
\caption{Analyses to quantify impacts of moral shifts (Section~\ref{subsec:analysis2}) . Higher scores indicate better performance.}
\label{fig:analysis2}
\end{figure}

While language use and moral values vary across social event domains, it is a separate question whether those variations will affect morality classification.
Prior work~\cite{hoover2020moral} has shown that classification performance varies across the social event domains when training and testing sets are from the same domain.
However, it is unknown how the domain variations impact the classification of moral values when the testing set is from a new domain.
To achieve this, we conduct a morality classification evaluation under the cross-domain setting that trains a classifier on one domain and tests the classifier on the other domains.
We split 80\% of documents as the training set and hold out 20\% of documents as the testing set for each domain corpus.
We extract TF-IDF weighted uni-, bi-, and tri-gram features on each domain corpus with the most frequent 15K features during the training.
We then build a logistic regression classifier using \texttt{LogisticRegression} from scikit-learn~\cite{pedregosa2011scikit} with default parameters.
Finally, we evaluate the classifier across each domain's test set using the F1 score.

We visualize cross-domain classification performance in Figure~\ref{fig:analysis2} with more details in the appendix's Figure~\ref{fig:cross_domain}.
In general, we can observe that in-domain classification evaluations outperform out-domain evaluations. 
For example, the in-domain evaluation on the Baltimore achieves 67.6\% versus the out-domain evaluation on the Election, 59.1\%.
The finding applies to the other cross-domain evaluations.
The observation indicates domain variations in moral values and language usage can impact morality classifier performance when training and testing sets of classifiers are from different social event domains.
To qualitatively examine the domain shifts of language use, we extract top word features with regard to morality labels for each domain and present a qualitative study in Appendix~\ref{subsec:top_feature}.

To statistically quantify the impacts of domain variations on cross-domain classification, we conduct a two-tailed t-test on the domain shifts.
The statistical test examines the domain and classification performance variations to test our null hypothesis that the domain shifts have no significant relation with the classification performance variations.
We first obtain performance variations by subtracting the in-domain score from the out-domain score. 
For example, to get the performance variation of Baltimore (in-domain) and Election (out-domain), we subtract the in-domain score (.676) by the out-domain score (.591).
To decide if the significant test uses parametric or non-parametric methods, we conduct a normality test via \texttt{scipy.stats.normaltest} from \texttt{statsmodel}~\cite{seabold2010statsmodels}.
The normality test yields p-value=0.533 for the domain variations and p-value=0.409 for the performance variations, which indicates that we will use parametric methods.
We then fit the two variables to a linear model ($Y_{performance} = f(X_{domain})+e$) that predicts performance variations by the domain variations (topic + label variations).
Finally, the test results show p-value=$1.21e-10$ and coefficient=$1.010$ for the correlation between the domain and classification performance variations.\footnote{p-value=$3.9e-5$ for the topic variation, and p-value=$3.43e-3$ for the moral label variation; coefficient=$1.012$ for the topic and coefficient=$1.008$ for the moral variation.}
Therefore, we reject the null hypothesis of the significant test.
The observations suggest that domain variations may cause significant variations for cross-domain classification performance.


\section{Learning to Adapt Framework (\textit{L2AF})}

\begin{figure*}[htp]
\centering
\includegraphics[width=0.824\textwidth]{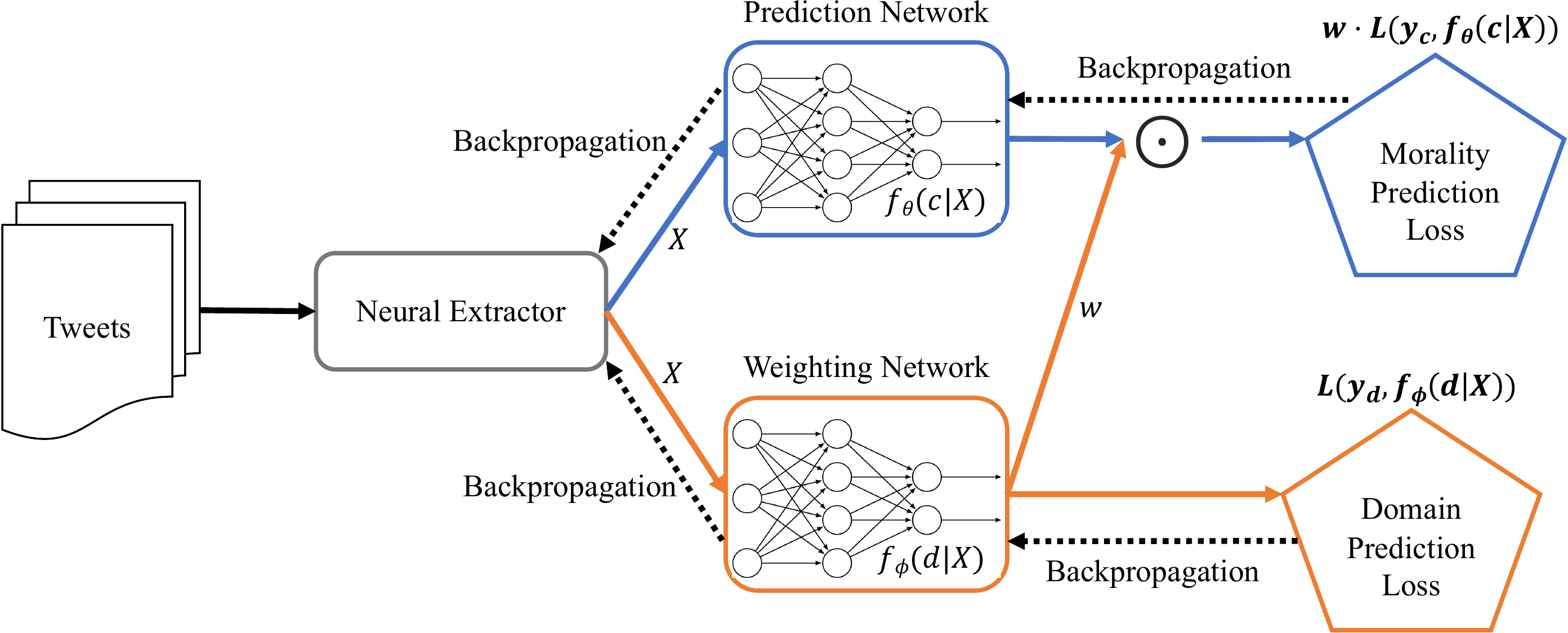}
\caption{Illustration of learning to adapt framework via instance weighting. We use blue and orange to represent optimization workflows of prediction and weighting networks respectively.}
\label{fig:model}
\end{figure*}

We propose a neural adaptation framework (\textit{L2AF} in Figure~\ref{fig:model}) that applies instance weighting~\cite{jiang2007instance} to encounter the domain variations for the morality classification task when training and testing sets are from different domains.
The framework adapts domain variations of language usage and moral values through four main modules: neural feature extractor, prediction network, weighting network, and joint optimization.

\subsection{Neural Feature Extractor}
Our framework treats neural models as feature extractors. 
The feature extractor encodes input documents into feature representations, $X$.
In this study, we experiment with two types of neural models, recurrent neural network (RNN)~\cite{cho2014properties} and BERT~\cite{devlin2019bert}, which have achieved promising performance in morality classification~\cite{johnson2018classification, hoover2020moral}.
The RNN is a bidirectional network with GRU~\cite{cho2014properties} units.
We then feed the document representations $X$ for two joint prediction tasks, prediction and weighting networks.

\subsection{Prediction Network}
The prediction network takes the document representations $X$ to predict moral values ($f_{\theta}(c|X)$) via a full-connected network, where $\theta$ is the network parameters and $c$ is the moral categories.
The network applies a dropout~\cite{nitish2014dropout} on the feature vectors $X$ and employs a softmax function to predict the 11 moral labels.
We apply the cross-entropy to calculate morality prediction loss and optimize the prediction network, $L(y_c, f_\theta(c|X))$, where $L$ is the loss and $y_c$ is the ground truth of moral values.

\subsection{Weighting Network}
The \textit{L2AF} deploys a weighting network that dynamically adapts the domain shifts of language and moral values.
Based on our observations (Section~\ref{subsec:analysis1}), we assume that two domains sharing similar language usage will also share similar moral values. 
And therefore, it is reasonable to assign higher weights to out-domain training instances sharing similar language usage with in-domain data and lower weights to out-domain training instances with different language usage from in-domain data.
The weighting network yields training weights ($w$) to leverage shifts between in-domain and out-domain data for the prediction network as $w \cdot L(y_c, f_\theta(c|X))$.
We obtain the weights by predicting in-domain probabilities of tweet documents, $f_\phi(d|X)$, where $d$ is domain label and $\phi$ refers to weight network parameters.
The $f_\phi(d|X)$ is a binary domain classification task that predicts the domains (in- vs. out-domain) of the tweet documents, where the $f$ is a sigmoid function to calculate in-domain probabilities, instance weights ($w$).
For example, if our target (test) domain is ALM (in-domain), we will treat the other six domains (e.g., BLM and Election) as out-domain.
We deploy a binary cross-entropy to optimize the weighting network.

\subsection{Joint Optimization}

\begin{equation}\label{eq:loss}
    L = \argmin_{\gamma, \phi, \theta} \alpha \cdot L(y_d, f_\phi(d|X)) + w \cdot L(y_c, f_\theta(c|X))
\end{equation}

We train the \textit{L2AF} via two optimization tasks (Equation~\ref{eq:loss}), moral value and domain predictions, where $\gamma$ refers to model parameters of the neural extractor, and $\alpha$ is a coefficient factor to leverage the importance of the domain prediction task. 
We first converge the weighting network during the training phase to obtain instance weights. 
We then train the prediction network with the instance weights. 
As the joint optimization continuously updates the neural extractor and weighting network parameters, the weighting network will dynamically adjust its predicted weights.
The dynamic procedure will ensure the framework adjusts and balance multi-domain data accordingly.
The framework trains the weighting network with Adam optimizer~\cite{kingma2014adam} and uses a separate optimizer for the prediction network.
If the neural extractor uses RNN, we optimize the prediction network with the RMSprop~\cite{tieleman2012lecture}. 
Otherwise, we optimize the prediction network with the AdamW~\cite{loshchilov2018decoupled} for the BERT-based extractor.
Separate optimizers for the joint prediction tasks can give us more controls on the model training process.
We only use the prediction network during the test phase to obtain moral value predictions.

\section{Experiments}

We focus on a multi-domain adaptation scenario in which morality classifiers trained on existing domains often adapt to a new domain quickly.
The goal is to evaluate the ability of a classifier to adapt from several observed domains to a new unobserved domain, which can be especially common for examining social movements by classifiers~\cite{mooijman2018moralization, reiter2021studying, stewart2021moving}.
We use the annotated morality data (Table~\ref{tab:moral_data}) that contains 30,979 tweets from 7 domains.
For each target domain, we randomly split 80\% of its domain data as a test set and the rest as a validation set for tuning model parameters to simulate limited annotated data from a new domain.
Under the multi-domain scenario, we train models on the 6 existing source domains, validate and tune model parameters on the validation set, and evaluate classification performance on the test set of the held-out target domain.
We report model performance by F1 score, which is a standard measurement in existing studies of morality classification~\cite{santos2019moral, hoover2020moral, rezapour2021incorporating}.

\subsection{Model Settings}
\label{subsec:settings}
We compare our neural framework (denote as \textit{Adapt}) with two baseline types, \textit{In-Domain} and \textit{No-adapt}.
The \textit{In-Domain} approach trains, validates, and tests classification models on in-domain (target domain) data.
The \textit{No-adapt} shares the same modules except for using the weighting network.
We train the \textit{No-adapt} approach by the same data splits with our proposed framework that trains classification models on out-domain (source domain) data, tune model parameters on the in-domain validation set, and evaluate models on the in-domain test set.
The \textit{No-adapt} approach follows the existing study~\cite{hoover2020moral} that predicts labels for multiple outcomes in a multitasking manner. 
We experiment with two types of neural classifiers, RNN and BERT.
For the base models (RNN and BERT), we utilize the pre-trained embeddings~\cite{pennington2014glove, devlin2019bert} to initialize model parameters.
The RNN and BERT baselines follow model architectures of existing studies, \cite{hoover2020moral} and \cite{mokhberian2020moral}, respectively, which have achieved state-of-art performance results in morality classification.
We trained the models on an Nvidia 3090 GPU and evaluated the model on CPUs.
Our experiments set the training batch as 16, the number of training epochs as 30, the max document as 60, and the dropout rate as 0.2.
We tune the learning rate in a range of [1e-6, 1e-4] on the validation set to get the best performance.

\subsection{Results}

\begin{table*}[ht]
\centering
\begin{tabular}{c|c|ccccccc}
\multirow{2}{*}{\makecell{Base \\ Model}} & \multirow{2}{*}{Type} & \multicolumn{7}{c}{Target Domain} \\
 &  & ALM & Baltimore & BLM & Davidson & Election & MeToo & Sandy \\\hline\hline
\multirow{3}{*}{RNN} & In-Domain & 0.700 & 0.633 & 0.705 & 0.918 & 0.551 & 0.549 & 0.518 \\
 & No-adapt & 0.716 & 0.674 & 0.711 & 0.932 & 0.653 & 0.553 & 0.526 \\
 & Adapt & 0.746 & 0.706 & 0.816 & \textbf{0.955} & 0.712 & 0.578 & 0.570 \\\hline\hline
\multirow{3}{*}{BERT} & In-Domain & 0.748 & 0.71 & 0.783 & 0.936 & 0.683 & 0.559 & 0.564 \\
 & No-adapt & 0.776 & 0.727 & 0.828 & 0.949 & 0.735 & 0.630 & 0.649 \\
 & Adapt & \textbf{0.781} & \textbf{0.732} & \textbf{0.864} & \textbf{0.955} & \textbf{0.758} & \textbf{0.655} & \textbf{0.672} \\ \hline\hline
\multicolumn{2}{c|}{$\Delta\uparrow$ (\%)} & 3.879 & 4.811 & 11.001 & 2.276 & 12.128 & 3.274 & 10.058 \\
\end{tabular}
\caption{Performance (F1 score) comparisons between three approaches across the domains of seven social movements. The $\Delta\uparrow$ refers to average improvements of our approach over baselines.}
\label{tab:performance}
\end{table*}

We report classification performance results in Table~\ref{tab:performance}.
Our learning to adapt framework leads to performance improvements (from 2.276\% to 12.128\%) over the comparable baselines across 7 domains.
The adaptation approach outperforms the \textit{No-adapt} approach by a large margin in both RNN and BERT models.
The improvements indicate that adapting domain shifts can improve morality classifiers and demonstrate that our approach can effectively adapt shifts in moral values and language usage across social events.
We find that our adaptation approach achieves minor improvements on the Davidson domain, and we infer this as the domain is more similar to other domains in moral values (Figure~\ref{fig:label_sims}) and language usage (Figure~\ref{fig:topic_sims}).
Training morality classifiers on multiple out-domain data can outperform the models trained on in-domain data.
Compared to the single-cross-domain evaluation (Figure~\ref{fig:cross_domain}), classifiers trained multi-domain data can perform better than in-domain-only data by leveraging common patterns across multiple domains.

\section{Case Study: Morality of COVID-19 Vaccine}

\begin{table*}[htp]
\centering
\resizebox{1\textwidth}{!}{
    \begin{tabular}{c|c|c||ccccc|ccccc|c||c}
    \multirow{2}{*}{Domain} & \multirow{2}{*}{Doc} & \multirow{2}{*}{Token} & \multicolumn{5}{c|}{Virtue} & \multicolumn{5}{c|}{Vice} &  & \multirow{2}{*}{\begin{tabular}[c]{@{}c@{}}Virtue-Vice\\ Ratio\end{tabular}} \\
     &  &  & authority & care & fairness & loyalty & purity & betrayal & cheating & degradation & harm & subversion & no-moral &  \\\hline\hline
    Vaccine & 500 & 18.59 & 7.88 & 3.11 & 0.37 & 5.86 & 0.55 & 3.30 & 2.93 & 6.78 & 9.16 & 3.85 & 56.23 & 0.683
    \end{tabular}
}
\caption{Data and annotation summary of COVID-19 vaccine.}
\label{tab:vaccine_data}
\end{table*}

In this section, we present a case study on the COVID-19 vaccine and examine if our \textit{L2AF} can effectively adapt morality classifiers trained on existing domains to a new target domain.
To verify our approach, we annotate Twitter data with morality labels following the definitions and annotation standards of the previous studies~\cite{haidt2007new, graham2013moral, hoover2020moral}.
We randomly sampled 500 vaccine-related tweets from a public repository~\cite{huang2020coronavirus} that has retrieved COVID-19 data using the Twitter streaming API since 2020.
We only keep English tweets and tweet authors from the US to reduce cultural and multilingual issues.
Two professional social psychologists (the second and third authors) participated in the data annotation process.
The annotators have demonstrated extensive expertise and published journal articles related to culture and morality~\cite{cohen2017psychology, northover2017effect, kwon2021changing}.
We utilized a double annotation strategy~\cite{dligach2011reducing}.
The first annotator reviewed each tweet and labeled the document with the 11 categories (10 moral values and one no-moral label).
The second annotator reviewed the annotations to ensure label qualities.
The two annotator setting fits this pilot study and reduces annotation biases beyond a single domain expert.
We follow the same steps in Section~\ref{sec:data} to preprocess the tweet documents and present the data summary in Table~\ref{tab:vaccine_data}.

\begin{figure*}[htp]
\centering
\includegraphics[width=0.774\textwidth]{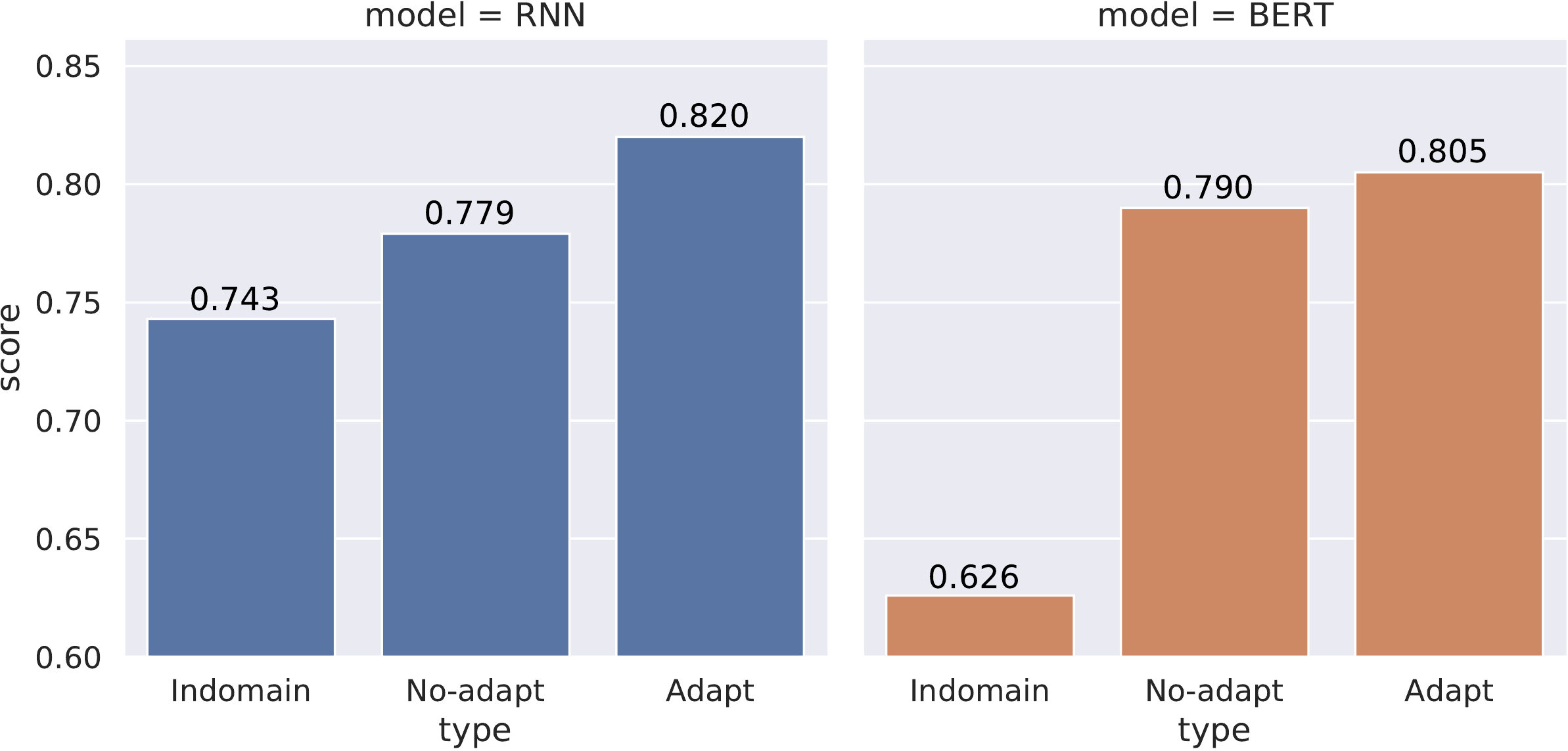}
\caption{Performance (F1 score) comparison on the COVID-19 vaccine data.}
\label{fig:vaccine}
\end{figure*}

We can find that the morality distribution is different from the existing 7 domains of social movements.
The table shows that: 1) authority and loyalty are the top two for virtue-related morality, 2) and degradation and harm are the top two for vice-related morality.
Such a finding aligns with a previous study~\cite{amin2017association} on parental vaccine hesitancy that traditional communications and interventions focusing on harm and fairness moralities may not fit for the new moral domain of vaccine hesitancy.
According to the Moral Foundation Theory~\cite{graham2013moral}, the corresponding moral values of authority, loyalty, harm, and degradation are subversion, betrayal, care, and purity.
The distribution differs from the previous study~\cite{amin2017association} on morality and vaccine hesitancy showing that moral values can also shift across the vaccine domain.
The domain shift of moral values highlights the necessity of developing generalizable and robust morality classifiers for understanding the vaccine challenge.

We follow the same experimental settings as Section~\ref{subsec:settings} and evaluate the three approach types on the COVID-19 vaccine data.
We present F1 scores for the two base models in Figure~\ref{fig:vaccine}.
Our approach (\textit{Adapt}) consistently outperforms the other two baselines across the two neural encoders.
The performance improvement demonstrates the effectiveness of our approach on adapting classifiers from existing source domains to a new target domain.
We also observe that the RNN-based classifiers generally outperform the BERT-based classifiers.
We infer this as the data annotation size that the BERT-based classifiers need more annotated data to achieve a similar performance than the RNN-based classifiers.
The adaptable and generalizable capabilities of classification models can be essential for studying the morality of new social movements when annotated data is rare or hard to obtain.

\section{Ethic and Privacy Concerns}

In this study, we only use the tweet documents and morality labels for evaluation purposes without any other user profile, such as user IDs.
All experimental information has been anonymized before training text classifiers.
Specifically, we used anonymized tweet IDs and replaced any user mentions and URLs with two generic symbols, ``USER'' and ``URL'', respectively.
To preserve user privacy, we will follow Twitter's privacy policy to release our annotated data of COVID-19 vaccine and provide instructions to access the public data~\cite{hoover2020moral} in our experiments.
We provided preprocessed datasets with the professional annotators to obtain high-quality annotations.
We report all preprocessing steps, hyperparameter settings, and other technical details of analysis procedures and will release our code repository to allow for replications.
Our quantitative and qualitative analyses rely on the machine learning toolkits, including topic model~\cite{rehurek2010software}, scikit-learn~\cite{pedregosa2011scikit}, and statsmodel~\cite{seabold2010statsmodels}.
Implementations of our approach and the baselines rely on the deep learning toolkits, including PyTorch~\cite{paszke2019pytorch} and Transformers~\cite{wolf2020transformers}.
Therefore the result report does not represent the authors' personal views.
All our observations derive from the reported results in this paper, which will be reproducible using our code.
Under Institutional Review Boards (IRB) guidelines, we believe that there is no code of ethics violation throughout the experiments and annotations.

\section{Related Work}

Emerging research studies have adopted the Moral Foundation Theory (MFT) to study moral values of users' behaviors and opinions among social issues, such as abortion~\cite{santos2019moral, rezapour2019enhancing, rezapour2021incorporating, roy2021identifying}, immigration~\cite{roy2021analysis, mokhberian2020moral, roy2021identifying}, climate change~\cite{brady2017emotion, rezapour2019enhancing, rezapour2021incorporating}, and natural disaster~\cite{garten2016morality, lin2018acquiring}.
Neural models (e.g., RNN and BERT) have dominated the classification performance of multiple natural language processing tasks~\cite{devlin2019bert}, text generation~\cite{emelin2021moral}, and question answering~\cite{lewis2020bart}.
Recent efforts have expanded the neural-based models to categorize moral values of online user-generated documents~\cite{lin2018acquiring, hoover2020moral, mokhberian2020moral}.

However, studies have found moral values can vary across domains of social issues~\cite{rezapour2019how, reiter2021studying} that may impact classification performance.
Current methods focus on augmenting document features for morality classifiers, such as incorporating background knowledge from Wikipedia~\cite{lin2018acquiring}, extending moral lexicon by crowdsourcing~\cite{hopp2021extended}, and vectorizing moral lexicons~\cite{araque2020moralstrength} to diversify token representations.
While the existing approaches focus on document feature augmentation, limited studies focus on the model level to explicitly incorporate the domain shifts into morality classifiers and adapt the classifiers to a new target domain.
Our work applies the instance weighting method~\cite{jiang2007instance, bhatt2016cross} and proposes a neural adaptation framework to augment the cross-domain performance of morality classifiers.
While previous study~\cite{kalimeri2019human} recruited participants by survey collections and identified moral values on vaccine attitudes from their liked Facebook pages, there is no prior study that examines moral values on COVID-19 from social media texts and develops classification models.
This study is the first work examine moral values in COVID-19 related tweets and has a great potential to probe hesitancy of COVID-19 vaccine from the social psychology aspects.

\section{Conclusion}
In this paper, we have examined domain shifts of language use and moral values across social issues, quantified the effects of the domain shifts on classification models, and proposed a neural adaptation framework to augment morality classifiers.
Differing from existing work of focusing on variations of moral values~\cite{rezapour2019how, reiter2021studying}, our study investigates both language use and morality shifts and analyzes connections across language use, morality, and classification performance.
Our approach dynamically adjusts weights on training instances from source domains regarding the new target domain and demonstrates its effectiveness on public data~\cite{hoover2020moral} and our annotated COVID-19 vaccine data.
To our best knowledge, the case study is the first pilot work investigating the moral values of the COVID-19 vaccine.
In our future work, we intend to extend the current morality annotations on the COVID-19 vaccine and explore shifts in other settings, such as geolocation and temporality.
Our code and data instructions will be available at \url{https://github.com/xiaoleihuang/MoralCausality}.

\subsection{Limitations}
While we have quantified of moral variations across social movements and proposed a neural adaptive framework to model the variations, we must acknowledge several limitations to appropriately interpret our findings.
First, we extracted language use patterns in Section~\ref{subsec:analysis1} via topic model, which are not accurate enough for short texts.
Using the derived topic distributions may contribute to the correlation test failure between language use and label variations.
Our future work will derive more robust feature representations for language use.
Second, the moral labels of the COVID-19 data by social psychologists may also bring annotation biases.
During the labeling process, one domain expert annotated the data and the second expert checked the annotations.
While annotation studies usually recruit at least three non-expert annotators by crowdsourcing methods (e.g., Amazon MTurk), recruiting multiple domain experts can be a challenge.
For example, the moral data~\cite{hoover2020moral} had multiple undergraduate students to assign moral labels to each tweet, which aims to reduce non-expert noisy and bias.
Combining experts and crowdsourcing annotators to build moral values of COVID-19 will be our future work.

\begin{acks}
The authors want to thank for reviewers' valuable comments.
\end{acks}

\bibliographystyle{ACM-Reference-Format}
\bibliography{sample-base}

\appendix

\section{Appendix Analysis}
\subsection{Cross-domain Performance Analysis}
\label{subsec:cdpa}

\begin{figure}[htp]
\centering
\includegraphics[width=0.904\linewidth]{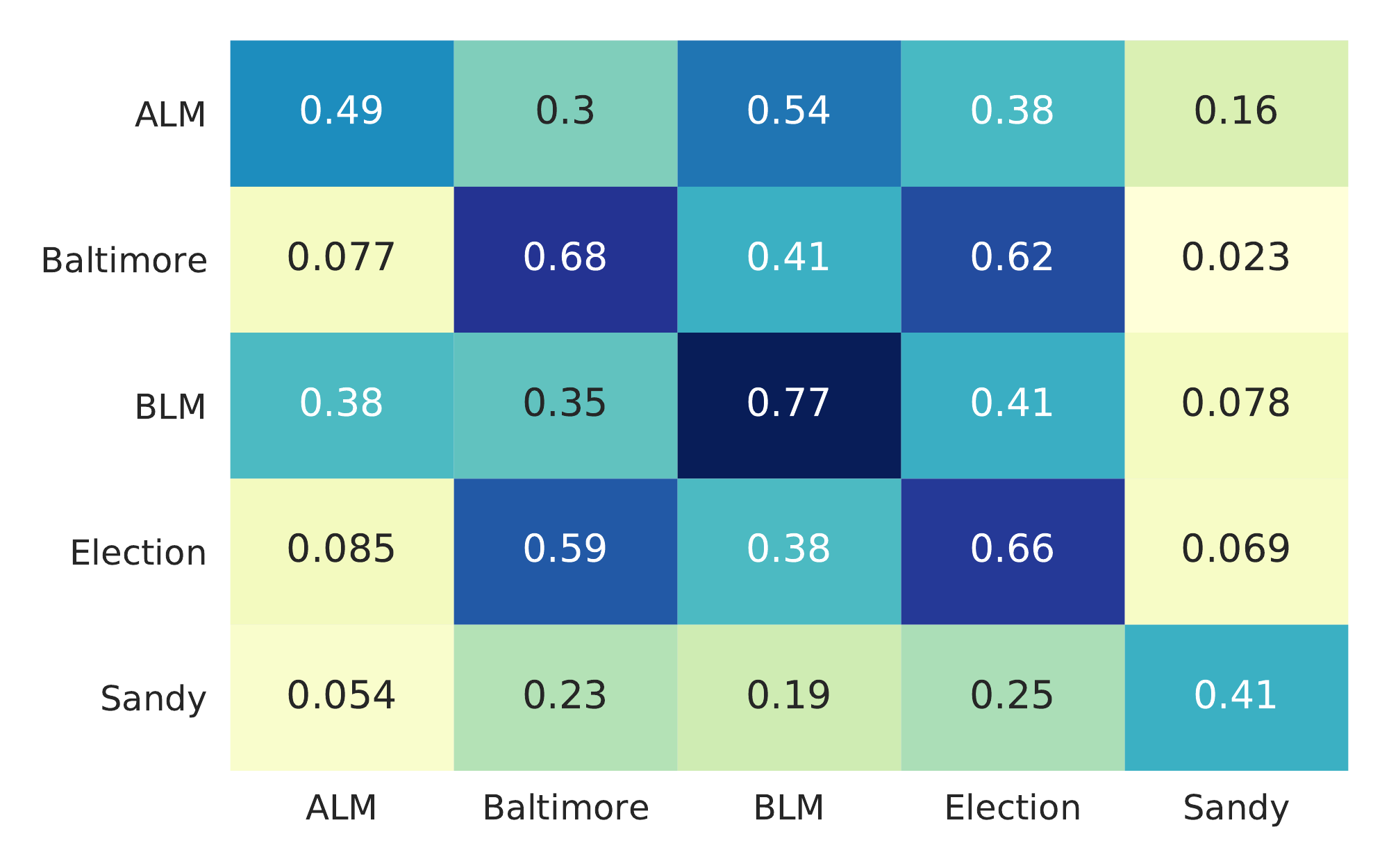}
\caption{Cross-domain classification performance by the F1 score.}
\label{fig:cross_domain}
\end{figure}

We report cross-domain performance of morality classifiers in Figure~\ref{fig:cross_domain}.
The cross-domain evaluations simulate the scenario that trains a classifier on existing source domain and applies the classifier on the new target domain.
The x-axis is for training domains, and the y-axis is for testing domains.
For example, given a pair of ALM (x-axis) and BLM (y-axis), the 0.38 indicates that we trained a classifier on ALM data and tested the classifier on the BLM domain.
The diagonal line indicates that we evaluate the classifiers within the same source domains.
We can find that evaluating classifiers within the same domain achieves better performance than out-domain evaluations.
We can also find that domains sharing similar topic features and moral values have closer performance.
For example, ALM and BLM are closer topics, and applying BLM-trained classifier on the ALM achieves 0.54, which is closer to the in-domain performance (0.77) than the other social issues.

\subsection{Top Feature Analysis}
\label{subsec:top_feature}

The quantitative analysis in Section~\ref{sec:data} and Appendix~\ref{subsec:cdpa} motivates us to explore further the domain shifts based on word features. 
To achieve this, we first binarize morality labels and build logistic regression classifiers using TF-IDF-weighted n-gram features (uni-, bi-, and tri-grams).
The classifier keeps the most frequent 15K features during the training.
We then train a classifier for each social issue domain and rank the word features by mutual information classification~\cite{pedregosa2011scikit}.
Finally, we remove stopwords by the NLTK~\cite{bird2004nltk} and present the top unigram features in Table~\ref{tab:features}. 

\begin{table*}[htp]
\centering
\caption{Top predictable unigram features regarding moral values extracted by the mutual information. The Vaccine refers to our annotated COVID-19 vaccine data.}
\begin{tabular}{c||c}
Domain & Features \\\hline\hline
ALM & people, god, justice, police, love, black, respect, human, racist, violence \\
Baltimore & freddiegray, baltimore, baltimoreriots, baltimoreuprising, police, people, justice, black, deray, man \\
BLM & blm, injustice, solidarity, police, black, people, ferguson, iuic, respect, blacktwitter \\
Davidson & bitch, hoes, bitches, pussy, fuck, shit, nigga, ass, lol, trash \\
Election & realdonaldtrump, trump, potus, president, justice, gop, maga, america, obey, donaldtrump \\
MeToo & god, love, people, justice, us, rights, human, hurt, respect, women, equality\\
Sandy & sandy, hurricanesandy, liberty, hurricane, holy, bitch, frankenstorm, god, love, obama \\
Vaccine & get, got, first, getting, vaccines, coronavirus, people, today, dose, worry
\end{tabular}
\label{tab:features}
\end{table*}

The qualitative results show the most predictable word features towards the 10 moral values.
Each domain shows its unique patterns of language use that reflect the morality, ideologies, and sentiments of online users.
We notice that the top features may reflect the societal and cultural backgrounds of the social issues.
For example, the top features of the Election domain correlate with political topics, such as maga (make America Great Again) and trump (US president); the Sandy happened in 2012 during Obama's presidency; and ``god'' connects to religious belief and ranks among top features in the multiple social issues.
We can observe that variations may also exist within the same social issue domain.
For example, while the hashtag ``baltimoreuprising'' supports the Baltimore protest, the hashtag ``baltimoreriots'' criticizes the protest as riots and violence.
The language use shifts suggest that such word feature variations may impact extracted feature representations and weaken morality classifiers for new target domains.

\end{document}